\documentclass[letterpaper, 10 pt, conference]{ieeeconf}  % Comment this line out if you need a4paper

\IEEEoverridecommandlockouts                              % This command is only needed if 
\usepackage{graphics} % for pdf, bitmapped graphics files
\usepackage{epsfig} % for postscript graphics files
\usepackage{mathptmx} % assumes new font selection scheme installed
\usepackage{times} % assumes new font selection scheme installed
\usepackage{amsmath} % assumes amsmath package installed
\usepackage{amssymb}  % assumes amsmath package installed
\usepackage{xcolor}
\usepackage{colortbl}
\usepackage{url}
\usepackage{gensymb}
\usepackage{algorithm}
\usepackage{algpseudocode}
\usepackage{listings}
\usepackage{caption}

\usepackage{hyperref}
\hypersetup{
    colorlinks=true,
    linkcolor=blue,
    filecolor=magenta,      
    urlcolor=blue
}
\def\BibTeX{{\rm B\kern-.05em{\sc i\kern-.025em b}\kern-.08em
    T\kern-.1667em\lower.7ex\hbox{E}\kern-.125emX}}
\usepackage{balance}

\newcommand{\commenting}[1]{}

\newcommand{\skill}[0]{\textit{Skill}}
\newcommand{\skills}[0]{\textit{Skills}}
\newcommand{\action}[0]{\textit{Action}}
\newcommand{\actions}[0]{\textit{Actions}}

\definecolor{Gray}{gray}{0.85}
\definecolor{LightCyan}{rgb}{0.88,1,1}
\definecolor{White}{rgb}{1,1,1}

\newcolumntype{g}{>{\columncolor{Gray}}c}

\makeatletter%change heading spacings
\renewcommand{\section}{\@startsection{section}{1}{\z@}{1.0ex plus 1.0ex minus 0.5ex}%
	{0.5ex plus 1ex minus 0ex}{\normalfont\normalsize\centering\scshape}}%
\renewcommand{\subsection}{\@startsection{subsection}{2}{\z@}{1.0ex plus 1.0ex minus 0.5ex}%
	{0.5ex plus 1ex minus 0ex}{\normalfont\normalsize\itshape}}%
\makeatother

\title{\LARGE \bf
Adaptation of Task Goal States from Prior Knowledge
}

\author{Andrei Costinescu and Darius Burschka$^{1}$% <-this % stops a space
\thanks{*This work was supported by the Lighthouse Initiative Geriatronics by StMWi Bayern (Project X, grant no. 5140951).}% <-this % stops a space
\thanks{$^{1}$All authors are with the School of Computation, Information and Technology at the Technical University of Munich. 
        {\tt\small \{andrei.costinescu, burschka\}@tum.de}}%
}

\begin{document}

\maketitle
\thispagestyle{empty}
\pagestyle{empty}

%%%%%%%%%%%%%%%%%%%%%%%%%%%%%%%%%%%%%%%%%%%%%%%%%%%%%%%%%%%%%%%%%%%%%%%%%%%%%%%%
\begin{abstract}

This paper presents a framework to define a task with freedom and variability in its goal state. A robot could use this to observe the execution of a task and target a different goal from the observed one; a goal that is still compatible with the task description but would be easier for the robot to execute. We define the model of an environment state and an environment \textbf{variation}, and present experiments on how to interactively create the \textbf{variation} from a single task demonstration and how to use this \textbf{variation} to create an execution plan for bringing any environment into the goal state.
\end{abstract}

%%%%%%%%%%%%%%%%%%%%%%%%%%%%%%%%%%%%%%%%%%%%%%%%%%%%%%%%%%%%%%%%%%%%%%%%%%%%%%%%
\section{INTRODUCTION}

The state of the scene that defines the goal of a specific task allows several possible variations in object positions and their internal state; variations that can still be considered valid configurations for a task. 

Many robotic systems try to accurately copy and imitate the goal configuration from human demonstration \cite{dobbe_immitationLearning, whirl_immitationLearning, mime_immitationLearning}. But they often struggle to do so because of additional challenges on the manipulator kinematics and, by imitating, they cost unnecessary additional time for adjustments of positions and internal states of objects, e.g. a cup's content level. Existing geometric task models \cite{taskLearning_trajectories} need to be extended by additional ontology information about possible variations in their execution, which this paper addresses.

% Fixed values with absolute precision are rarely present in nature. A \underline{range} of temperature values in which water is liquid favored the emergence of life. There is a \underline{period} during the year when farmers plant seeds for new plants to grow, not a single second when they must be planted. Even humans have a \underline{range} of hunger levels that we can tolerate before we go and eat.

% Furthermore, we, humans, do not have accurate sensors: we can not pinpoint exactly how many millimeters apart two cups are in the cupboard, can not precisely determine the 3d location of a sound source by hearing it, or determine the exact temperature of a hot cup by touching it. Thus, we have come to live by ranges of values, not fixed ones.

In non-assembly tasks, e.g., household environments, it is rare that fixed values are desired for object states. 
When cleaning the table after dinner, the washed cutlery does not have a fixed 3d pose in the cutlery drawer. Rather, there is an assigned space, a zone, a range of values where the knives, forks, and spoons are located, but not fixed values. 
% Similarly, when requesting a cup of water, we do not think of the exact amount of water volume to be poured into the cup. We have an accepted threshold of the content level when we consider a cup to be full. 
Similarly, when we want to empty a cup, i.e. set its content to 0, we do not make sure there is no water atom left in the cup.

Thus, representing task goal states as (only) fixed values does not model reality. 

\begin{figure}[t!]
    %\vspace{-0.5cm}
    \centering
    \includegraphics[width=1\linewidth]{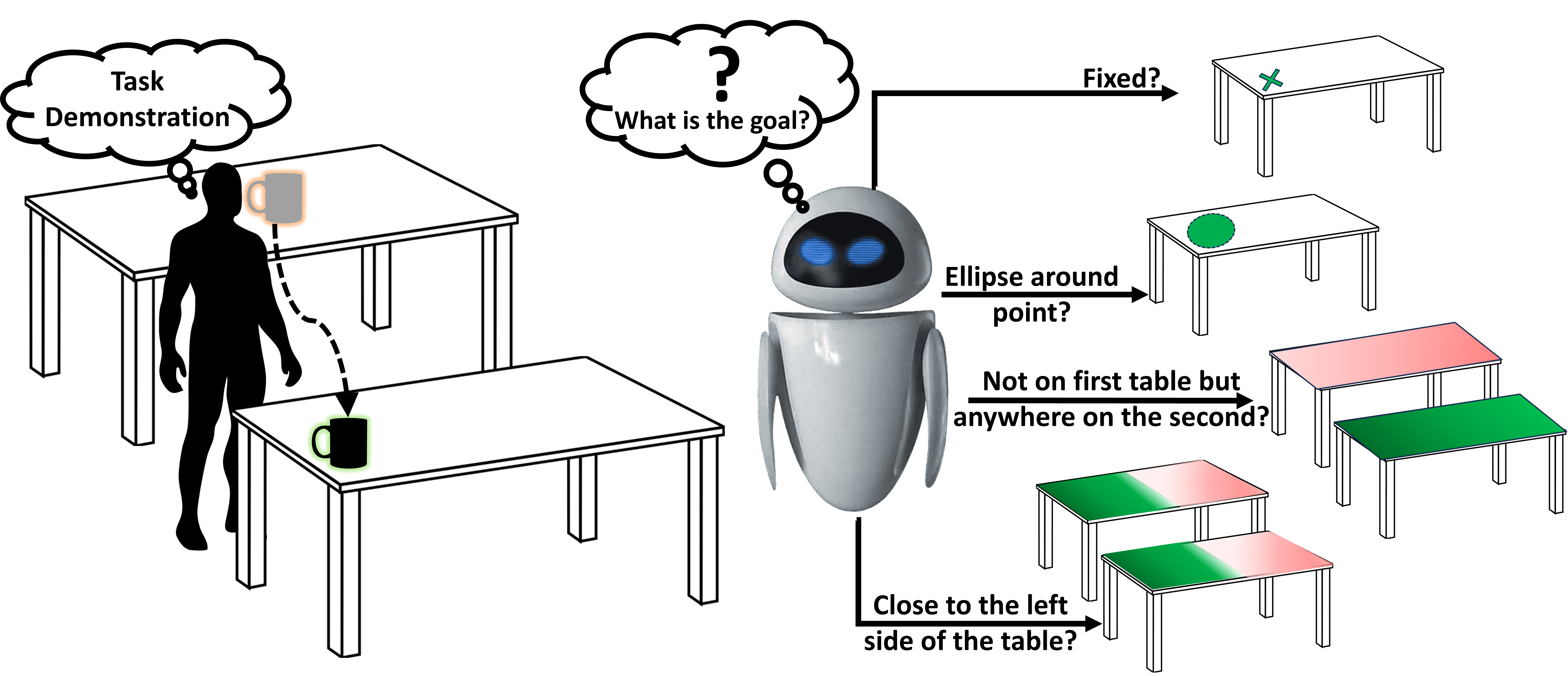}
    \caption{Understanding, i.e. representing, the intended amount and type of variation in a task's goal state is paramount for task monitoring and execution planning.} \label{fig:teaser}
\end{figure}

The option to model ranges of values is important for robots to make sense of the world: to learn task goal states from the persons with whom they interact and to execute skills to bring an environment into a particular goal state. This is a challenging problem, as illustrated by Figure \ref{fig:teaser}. The user, moving the cup on the front table, has a goal definition in mind that he is trying to achieve. As illustrated on the figure's right side, there are many possibilities for the intended goal. A method of disambiguation and navigation through the possibility space is required for task goal specification.

In this paper, we propose a way of representing task goal states not via fixed states but via value ranges of agent and object properties. This is based on a hierarchical, conceptual definition of objects and agents that includes (both intrinsic and extrinsic) properties with their respective co-domain, which we call \textit{ValueDomain} \cite{conceptHierarchyGeriatronicsSummit24}.

Following the structured definition of environment states, we define a task’s goal state as a subset of the whole environment \textit{ValueDomain}. We call a subset of a \textit{ValueDomain} a \textbf{variation}. This structured environment definition enables computing differences between values of the same \textit{ValueDomain} and between values and defined \textbf{variation}. This becomes important when trying to solve a task: i.e. determining the \skills\ for agents to execute to change the environment into the task’s goal state. Creating the \textbf{variation} of an environment is also important to the model. We propose an interactive method based on one user demonstration of a task. We create the task goal state by computing the differences between the pre- and post-demonstration environments and by asking the user questions to solve ambiguities in the \textbf{variation} selection process.

Task goal modeling was developed as part of classical artificial intelligence (AI) tools. Initially, logical predicates and statements were used to define states and (fixed) goal conditions. This definition led to the development of constraint satisfaction problems \cite{ai_book_1}, in which variables could be constrained to ranges of values, not only fixed ones. The possible value ranges were also extended from Boolean to numerical values.

Version Spaces \cite{versionSpaceAlgebra} is another classic AI tool for modeling variable states. Though initially not developed for modeling task goals, Version Spaces represent a set of hypotheses explaining a supervised data set. It can represent a most general hypothesis, accepting any data, a most specific hypothesis, accepting only one data point, and all other possible combinations of data points. One can view Version Spaces as representing subsets of a set: the set itself, single-element sets, and all element combinations in between. Inspired by this set-like view on Version Spaces, we have developed the \textbf{variations} of \textit{ValueDomains}.

STRIPS \cite{strips} emerged as a way of representing changes to states, called actions, that have preconditions to be satisfied and effects on the entities on which the action is performed. PDDL \cite{pddl} is a language in which many planning problems are specified. In it, a domain, i.e. general knowledge about the modeled world, a problem, i.e. the task goal, and an initial state are defined. In the problem domain, an enumeration of actions is included; these actions are sequenced by solvers to reach the task goal defined in the problem. PDDL supports many logical and temporal constraints and also defining variable goal states. However, more complex data types, such as lists or maps like one would need to maintain the collection of entities in an environment, are not supported. A distinction between actions and skills was proposed in \cite{conceptHierarchyGeriatronicsSummit24}, in which actions represent the abstract change to be executed and skills, performed by agents in environments, execute the change represented by actions in the real world.

Recent approaches no longer model tasks but employ large language models (LLM) to create task plan execution from a textual task description \cite{taskModel_llm1, fltrnn, taskModel_llm2}. Most tasks specified in this format already include the steps, i.e. actions, for a robot to perform: e.g. "\underline{put} the mug in box" or "\underline{open} the drawer" from \cite{taskModel_llm1} or "\underline{open} bathroom window" from \cite{taskModel_llm2}. Also, in \cite{fltrnn}, the goal state of a task is fixed, not variable. The authors of \cite{task_onlyPose_withVariableGoal} propose a method for solving object arrangement tasks. While not modeling internal object states, just 3d poses, the tasks have non-fixed goal states, as expected from e.g. household dinner setups.

Task learning from observation approaches include behavioral cloning \cite{bco}, learning from demonstrations \cite{learningFromDemonstrations}, and interactive learning \cite{interactiveLearning}. We employ an interactive learning from demonstration method to construct our task goal model from one visual task demonstration.

KnowRob \cite{knowrob} and ConceptNet \cite{conceptNet} are ontologies, i.e. knowledge bases, to store knowledge about robot skills and natural language, respectively. We use the concept hierarchy presented in \cite{conceptHierarchyGeriatronicsSummit24} to model \textit{ValueDomains}, objects, agents, actions, skills, and their properties.

\section{MODEL}
Task goal representations in household environments can specify fixed values for entity properties, a range of allowed values, or be indifferent to the property value. In the example of Figure \ref{fig:teaser}, the goal state could be that the drinking mug must be half full, and its accepted location should be close to its current location on the front table. In this goal example, there is a fixed value for the amount of content of the drinking mug, a range of values for its location, and any possible value for the table's location, because it was not specified in the goal state. A model of a task's goal must represent all these possibilities. Our approach is to introduce \textbf{variations} of values, described in \ref{ssec:variations}.

To turn an environment into a desired goal state, the system must first assess the current \underline{state} of an environment, described in \ref{ssec:environment_state}. In the environment state, values of entity properties and the whole range of allowed values are represented. This range of allowed values is the basis of \textbf{variations}, which can be seen as subsets of the whole domain of values.

In the second step, the differences between the current and goal environment states must be computed, see \ref{ssec:comparisons}, and solved to get to the desired task goal state.

\begin{figure}[t!]
    %\vspace{-0.5cm}
    \centering
    \includegraphics[width=1\linewidth]{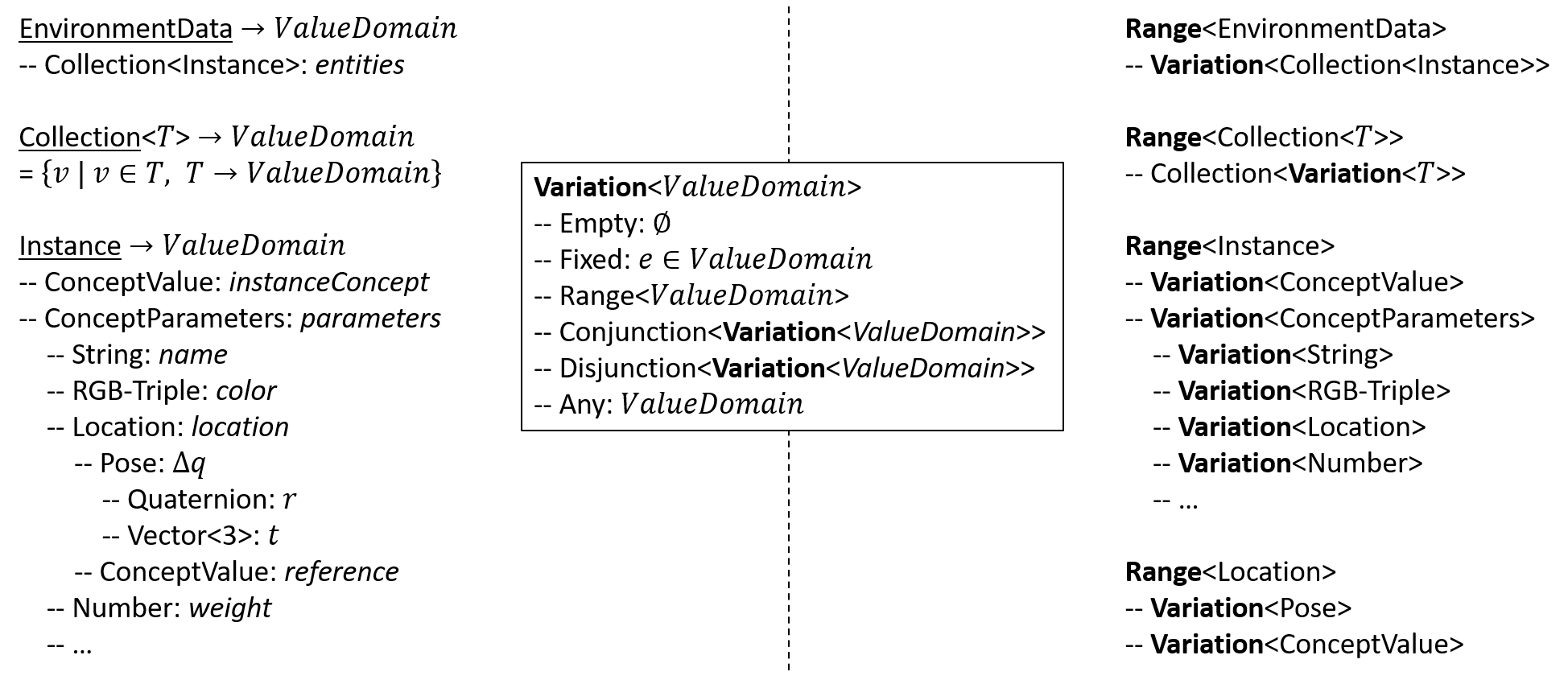}
    \caption{Variations can be used to express a desired range of values for environment states. On the left are the contents of an environment state. The right part shows RangeVariations of the \textit{ValueDomains} on the left. $A \rightarrow B$ means $A$ is a subtype/-concept of $B$. \textbf{Bold types} are variations.} \label{fig:objects_and_tasks_domain}
\end{figure}

\subsection{Model of the world: What is an Environment State?}\label{ssec:environment_state}
We model the state of an environment as the collection of states of entity instances (i.e. agents and objects) in the environment, see the left side of Figure \ref{fig:objects_and_tasks_domain}. In the environment on the left side of Figure \ref{fig:teaser}, there are four entity instances: the person doing the demonstration, the drinking mug that has changed location, and the two tables, one in the back and the other in the front, on which the mug is now located.

The state of an entity instance comprises the concepts of the entity and the collection of properties defined by the entity's concepts \cite{conceptHierarchyGeriatronicsSummit24}. Concepts also define their properties' allowed range of values (incl. value type). Subconcepts inherit properties. Instances, not concepts, define the values of their properties. The state of the drinking mug thus comprises all parent concepts, e.g. mug, liquid container, container, object, physical entity, concept, and all properties defined by these concepts, including the location, mass, maximal volume that can be contained (\textit{contentVolume}), volume currently contained (\textit{contentLevel}), contained instances, color, etc.

Thus, the environment state contains all the property values of its entity instances.

The property values are elements of a \textbf{\textit{ValueDomain}}, i.e. a set of values. For example, the set of values of the \textit{contentVolume} property is a non-negative real number, and the set of values of the \textit{location} property is a tuple of the reference entity and the pose delta to that reference's origin. To represent \underline{subsets} of \textit{ValueDomains}, we define \textbf{variations}.

\subsection{What is a Variation?}\label{ssec:variations}
A \textbf{variation} of a \textit{ValueDomain} represents a subset of values from that \textit{ValueDomain}. Thus, a variation can be empty, it can be a fixed value, it can be a range of values (\textbf{RangeVariation}), it can be a conjunction and/or a disjunction of variations, or it can be the whole \textit{ValueDomain} itself.

For example, the variation of an \textit{Integer} value could be the empty set, the number $4$, the set of prime numbers, the union of the integer intervals $\left[2; 5\right]$ and $\left[36; 42\right]$, the intersection of the integer intervals $\left[2; 5\right]$ and $\left[4; 9\right]$, or $\mathbb{Z}$-itself.

Similarly, the variation of a \textit{ConceptValue} could be the empty set, the concept \textit{Object}, the \textit{Object} concept including all its subconcepts, the \textit{RedObject} and/or \textit{GreenObject} concepts, or the set of all concepts.

An example of a meaningful RangeVariation of an entity \textit{Instance} is the variation of the entity's \textit{ConceptValue} and the variation of all properties that are defined by the \textit{ConceptValue} variation.

A meaningful RangeVariation of a collection, e.g. the collection of entity instances in an environment, is a set of variations for the collection's elements. Not all collection elements must satisfy a variation, but all variations must be satisfied by (different) elements. For referencing, we name this variation type as $A \equiv \{v \in \text{\textbf{Variation}}\left<\text{\textit{CollectionType}}\right>\}$. Given $x \in \text{Collection}\left<\text{CollectionType}\right>$, $x \in A \Leftrightarrow \forall v_e \in A,$\\$\exists e \in x: e \in v_e$.

Finally, a meaningful RangeVariation of an environment state is the variation of the collection of entity instances, see the right side of Figure \ref{fig:objects_and_tasks_domain}.

\subsection{Comparison between ValueDomains}\label{ssec:comparisons}
In our work, a \textbf{Comparison} represents the contrast between two \textit{ValueDomains}. It contains a target and a value to be checked for equality with the target.

Values and targets of different \textit{ValueDomains} are different. Values and targets of the same ValueDomain are compared for equality. If different and the \textit{ValueDomain} has sub-data, Comparisons of the sub-data are also created. For example, as per Figure \ref{fig:objects_and_tasks_domain}, a \textit{Location} is represented as a \textit{Pose} delta to a reference \textit{Instance} entity. Thus, the \textit{Location} has a \textit{Pose} and an \textit{Instance} as sub-data. If the \textit{Location} target and value are different, Comparisons of the \textit{Pose} and \textit{Instances} are also created. This helps pinpoint the exact reason for the values being different. Similarly, \textit{EnvironmentData} has a \textit{Collection} of \textit{Instances} as sub-data. \textit{Collections} have an \textit{Integer} size and elements as sub-data. 
\textit{Instances} have a \textit{ConceptValue} and \textit{ConceptParameters} as sub-data.

Comparisons of sub-data are saved as additional information in the parent data Comparison so that, e.g., the \textit{Location} Comparison knows there is a difference between the \textit{Pose} sub-data of the \textit{Location}.

Our system can also model Comparisons between a \textit{ValueDomain} and a \textbf{variation} of the same \textit{ValueDomain}. In this case, the Comparison is equivalent to checking if the value is inside the \textbf{variation}. If not, the additional information stored in the comparison are the reasons why the value is not in the target \textbf{variation}. In the example of an \textit{Integer}-Comparison, if the target \textbf{variation} is the interval $\left[6, 10\right]$ and the value is $4$, the reason is the \textit{Boolean}-Comparison of the \textit{LessEqual} function between the interval's lower bound $6$ and the value $4$, which is false but is supposed to be true when the value is inside the interval.

\begin{figure}[t!]
    %\vspace{-0.5cm}
    \centering
    \includegraphics[width=1\linewidth]{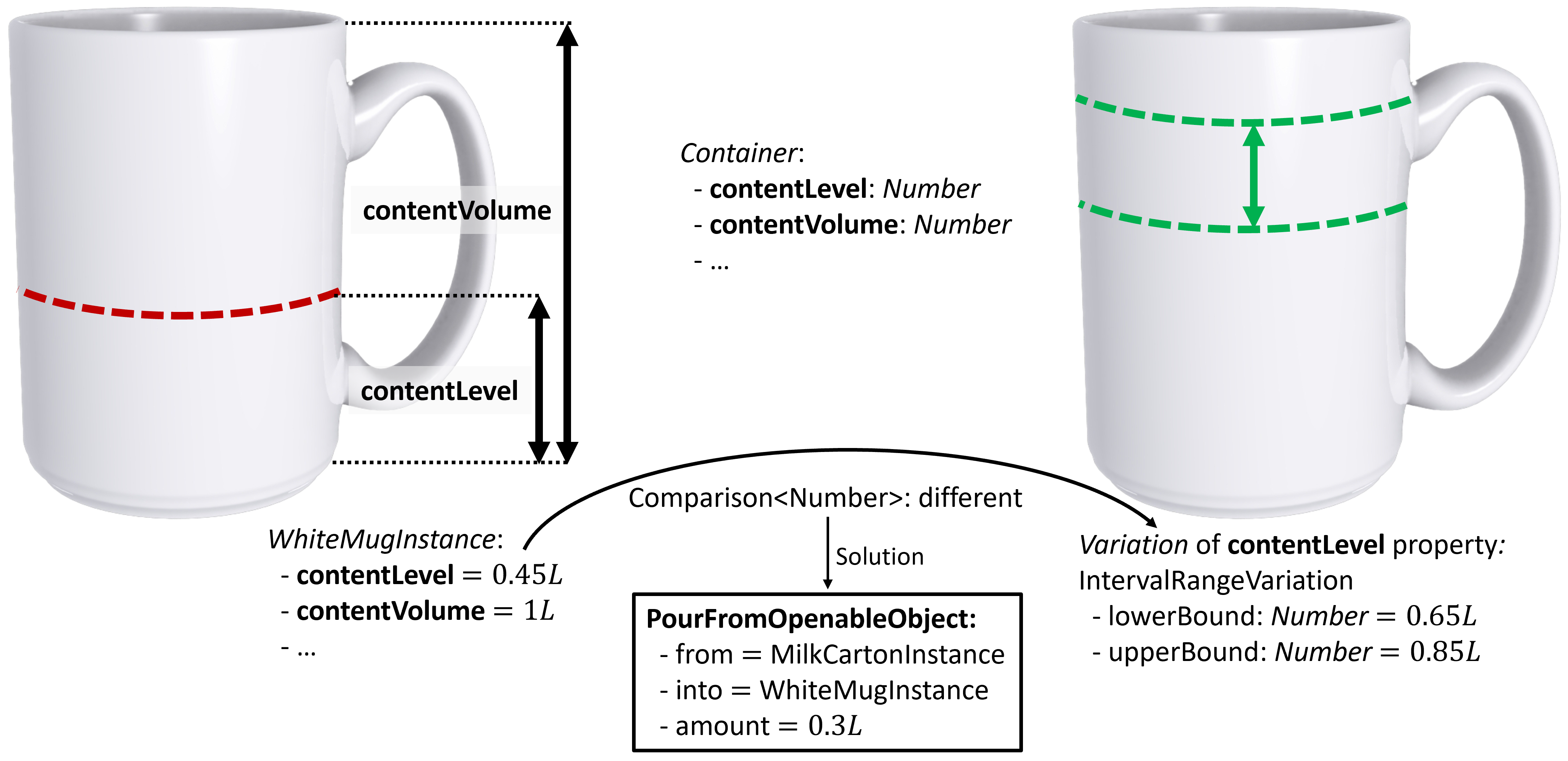}
    \caption{The \textit{Container} concept defines the \textit{contentLevel} property as a \textit{Number}. This property has a value of $0.45$ in the \textit{WhiteMugInstance}. One (or more) \skill(s) must be executed to bring the current level to the desired level inside the defined variation on the right.} \label{fig:concept_property_difference}
\end{figure}

The third Comparison type is when the value stems from an \textit{Instance}'s concept properties. Besides the additional info when the value is different from the target, this Comparison type has information about the instance, which the system can use to determine which \actions\ and \skills\ (see \ref{ssec:actions_skills}) should be used to change the value of this concept property. Figure \ref{fig:concept_property_difference} illustrates a solution to bring the \textit{contentLevel} property of the \textit{WhiteMugInstance} inside the variation range.

\section{RESULTS}

\begin{figure*}[t!]
    %\vspace{-0.5cm}
    \centering
    \includegraphics[width=1\linewidth]{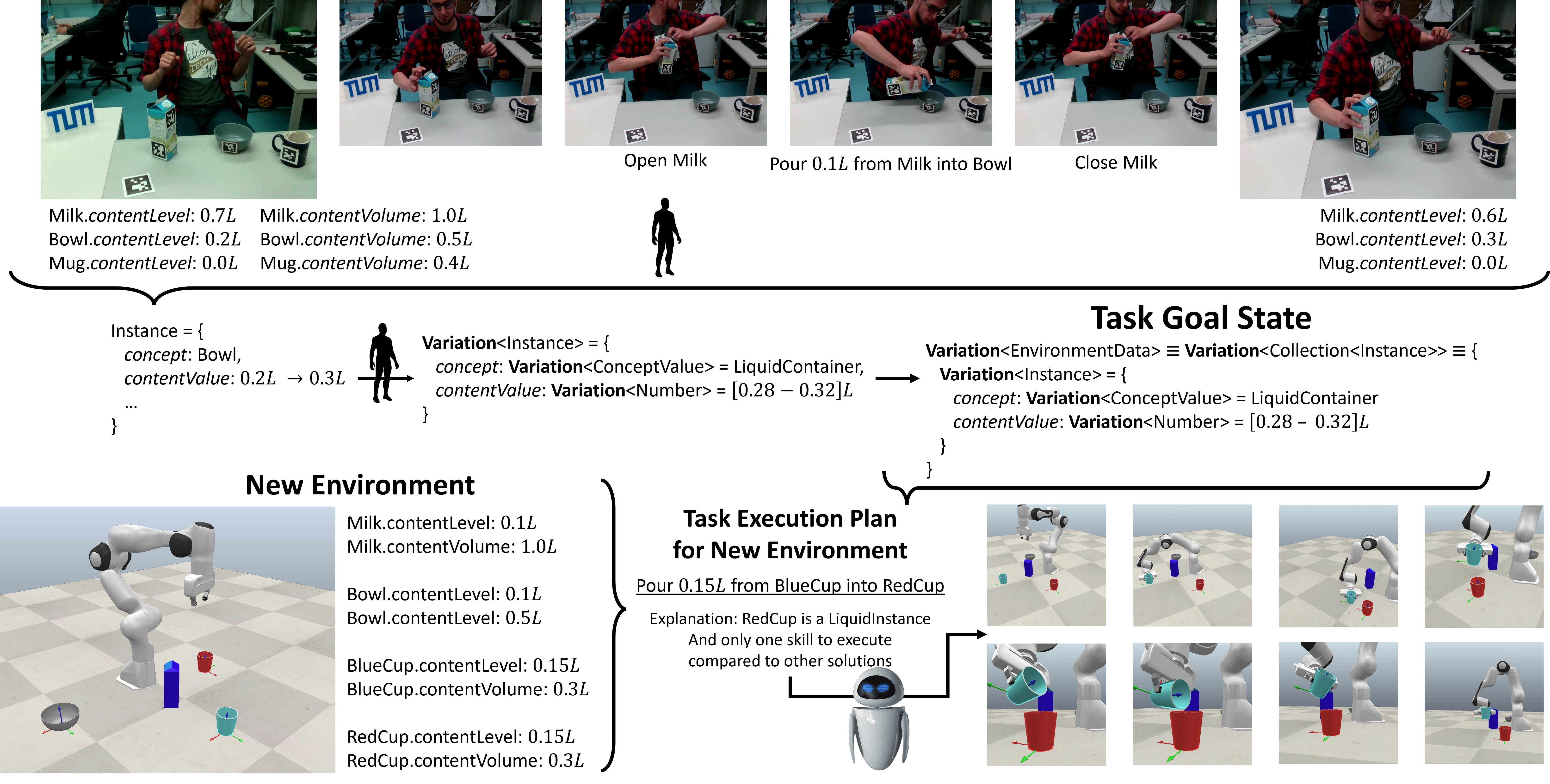}
    \caption{From a single user demonstration, the system extracts the desired task goal state with the help of user interaction to solve ambiguities. Using the created environment variation, the system computes a task execution plan to bring new environments into the goal state. It sends the plan to agents in the environment to execute.} \label{fig:system_architecture}
\end{figure*}

Figure \ref{fig:system_architecture} shows our proposed framework to define a task goal, i.e. an environment goals state, and to turn a given environment into this goal state. The system visually observes a task execution by a user and segments this \underline{single} demonstration into \skills. \actions\ and \skills\ are defined in \ref{ssec:actions_skills}. The demonstration changed one or several properties of entities in the environment; environment which is now in the goal state. This information and the differences in entity properties from the start and end environment states are used to represent the task goal state. More on that in \ref{ssec:exp_model_def}. To turn a new environment into the defined goal state, a planning problem must be solved. This entails computing the differences between the environment's current state and the goal state, finding \actions\ that solve these differences, instantiating \skills\ that implement the \actions\ in the environment, selecting the \skills\ to execute by minimizing a given metric, and finally, sending the \skills\ to the agents in the environment to execute. This process is detailed in \ref{ssec:exp_model_use}.

% To prove the usability of our model, we present experiments to create a new goal state and turn the current environment into an (already-defined) goal state.

\subsection{Actions and Skills}\label{ssec:actions_skills}
A change in the environment is modeled using \actions, i.e. \textbf{what} has happened, and \skills, i.e. \textbf{how} did the change happen \cite{conceptHierarchyGeriatronicsSummit24}. Like in STRIPS \cite{strips} and PDDL \cite{pddl}, we represent \actions\ by their effects on entity properties and \skills\ by their preconditions and effects. \actions\ do not need preconditions because they only describe the \textbf{what} part of a change, not which conditions must be satisfied to perform the change. Besides preconditions and effects, \skills\ have a list of checks that tell our system if the \skill\ is executed in the environment. These checks allow the creation of a \skill\ recognition program, like the one presented in \cite{conceptHierarchyGeriatronicsSummit24}.
%\todo{citation of Geriatronics summit paper or the journal/unsubmitted paper?}
Using the \skill\ recognition output, we capture the changes from a task demonstration.

A \skill\ is thus the physical enactment of an abstract \action\ in an environment. Hence, \skills\ are correlated with \actions\ via their effects. A \skill\ can have more effects than a corresponding \action. For example, the \skill\ of scooping jam from a jar with a spoon implements the \action\ of \textit{TransferringContents}, but it also \textit{Dirties} the spoon.

\subsection{How To Parameterize The Model}\label{ssec:exp_model_def}
Creating a new goal state should be easier than manually specifying all variations wanted from the goal state. Doing so requires programming knowledge, which should not be needed to define goal states. One can let the system, which knows how to represent goal states, question the user about the desired state of the environment. However, this tedious process requires many questions from the system, also leading to decreased system usability.

Therefore, our approach is to let the user turn a given environment into a desired goal state and analyze the differences between the initial and final environment state to create the goal state representation. This single demonstration highlights the entity property values that were not in the desired goal state before being changed by the user.

We capture the demonstration via an Intel Realsense 3D camera \cite{realsense}, analyze the human skeleton via the OpenPose human pose estimation method \cite{openpose}, and determine the 3d pose of objects with AprilTag markers \cite{aprilTag}.

One demonstration contains the initial environment, not in the task goal state, and the final environment, in the goal state. The final environment state alone is not enough to create the environment variation. Thus, additional questions, guided by the differences between the two environment values, are posed by the system to the user to determine the desired variation in the environment state.

In a demonstration in which the user pours milk into a bowl, as shown in the top of Figure \ref{fig:system_architecture}, the initial question posed to the user is which entities that have changed properties are relevant for the goal state. If the goal state is to have more milk in the bowl, the milk carton is irrelevant; it is a means to achieve the goal state but not relevant to the goal itself. The bowl is thus selected as a relevant entity. 

Next, the list of relevant modified properties must also be determined for each relevant entity. It could have happened that during pouring of the milk into the bowl, the bowl's location also changed, e.g. touched accidentally by the user. Thus, not all modified properties could be relevant to the task. After selecting the relevant properties, the system knows from the knowledge base \cite{conceptHierarchyGeriatronicsSummit24} their \textit{ValueDomain} and the list of implemented \textbf{variations} for that \textit{ValueDomain}. Thus, the user parametrizes a selected \textbf{variation} from the list: choosing either a fixed value, a \textit{ValueDomain}-specific \textbf{RangeVariation} that must be parametrized, a conjunction or disjunction of \textbf{RangeVariations}, or the whole \textit{ValueDomain}.

In the example above, the user chooses the \textit{contentLevel} property as relevant. The system knows this property's defined set of values: a non-negative real number, and the possible range variation types: an open interval, a closed interval, an open-closed or closed-open interval, an intersection or union of intervals, etc. The user chooses a closed interval of $[0.28, 0.32]$ around the final \textit{contentLevel} value of $0.3L$. The user also specifies a variation for the entity's concept. It is generalized from that specific bowl instance to a \textit{LiquidContainer}.

After each modified property of each entity has a represented \textbf{variation}, the system automatically collects the entities into a variation of type $A$, see \ref{ssec:variations}, which is the assigned \textbf{variation} for the collection of entities in the environment.

\begin{figure}[t!]
    %\vspace{-0.5cm}
    \centering
    \includegraphics[width=1\linewidth]{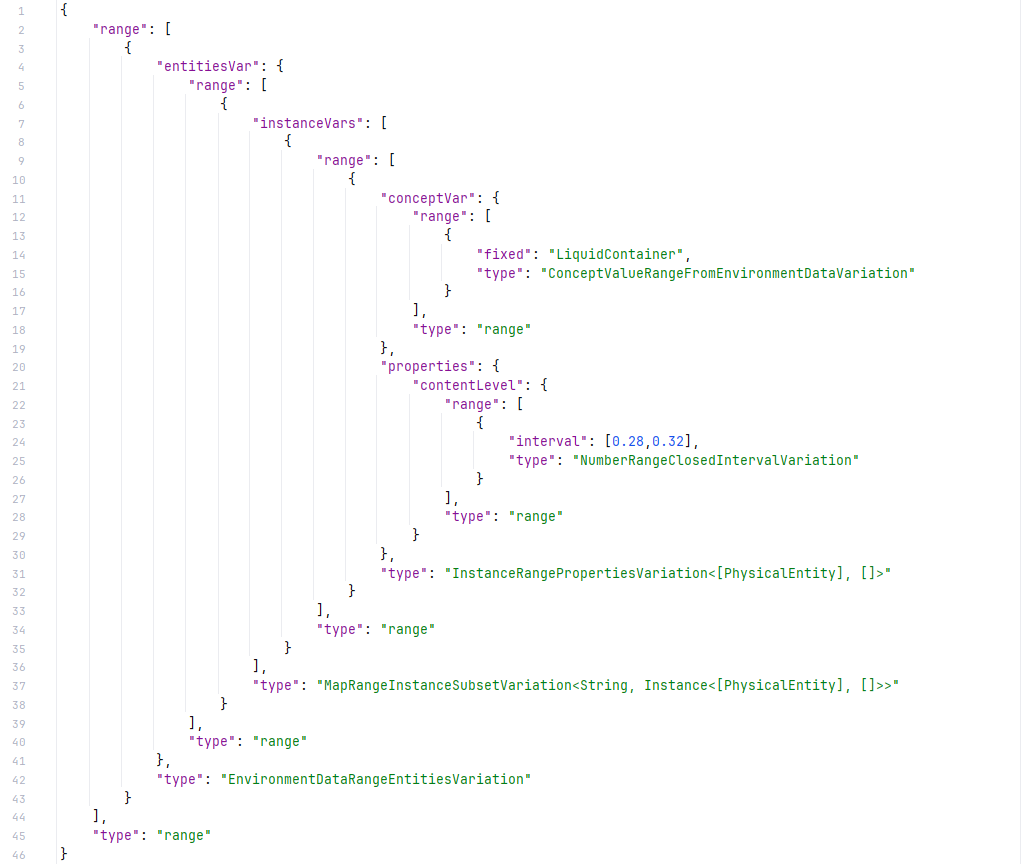}
    \caption{The goal state is a \textbf{RangeVariation} of the environment, of type EnvironmentDataRangeEntityVariation, which contains a \textbf{variation} of entities. This sub-variation is a \textbf{RangeVariation} of type MapRangeInstanceSubset (\textbf{variation} of type $A$, see \ref{ssec:variations}) and contains one instance \textbf{RangeVariation} of type InstanceRangePropertiesVariation. It defines the instance's concept \textbf{RangeVariation}, a \textit{LiquidContainer} to be found in the environment, and the \textit{contentLevel} property \textbf{RangeVariation}, the closed interval $\left[0.28, 0.32\right]$.} \label{fig:task_goal_state}
\end{figure}

Thus, the environment variation is determined in $\mathcal{O}\left(n\times m \times p\right)$ questions to the user, where $n$ is the number of entities in the environment, $m$ is the maximal number of properties that an entity can have, and $p$ is the maximal number of parameters that a \textbf{RangeVariation} needs to be represented. In the example above, $10$ questions were necessary to determine the task goal state shown in Figure \ref{fig:system_architecture} of a \textit{LiquidContainer} with \textit{contentLevel} between $0.28$ and $0.32L$. Figure \ref{fig:task_goal_state} shows the internal JSON-like representation of the goal state as the environment variation.
% 1 question which entities are relevant -> just bowl
% 1 question which properties are relevant -> contentLevel and concept
% 1 question about concept values being the same; should create variation?
% 1 question which ConceptValue-variation to select -> ConceptValue in Environment
% 1 question: which generalized concept?
% 1 question -> add other range-variation
% 1 question which Number-variation to select -> Interval
% 1 question: min-bound?
% 1 question: max-bound?
% 1 question -> add other range-variation

\subsection{How To Use The Model}\label{ssec:exp_model_use}
\begin{figure}[t!]
    %\vspace{-0.5cm}
    \centering
    \includegraphics[width=1\linewidth]{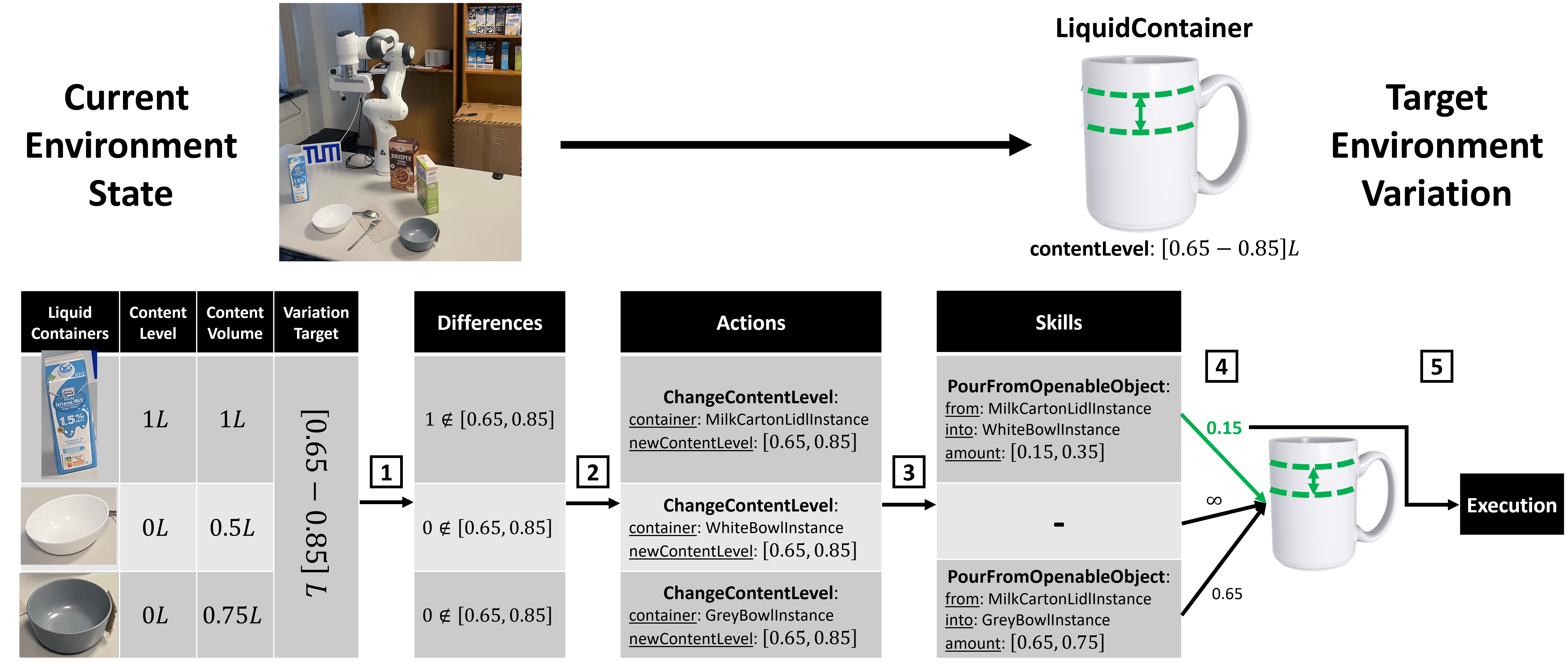}
    \caption{The procedure to turn an environment into its goal state is divided into 5 steps: computing differences, finding abstract solutions (i.e. \actions), computing practical solutions for the abstract ones (i.e. \actions\ $\rightarrow$ \skills), selecting the best practical solution, and executing the solution.} \label{fig:experiment_description}
\end{figure}

Assuming the representation of a task's goal state is given, i.e. an environment variation, we detail our procedure (see Figure \ref{fig:experiment_description}) to turn the current environment into the goal state.

First, a Comparison between the environment and the goal variation is computed. This leads, as described in \ref{ssec:comparisons}, to a list of reasons why the environment is not in the variation. These reasons, i.e. differences $\delta$ of concept properties $p$, must be fixed to turn the environment into the goal state.

% Computing the differences between an EnvironmentData and an EnvironmentData-Variation, that has a Collection-Variation of type $A$, see \ref{ssec:variations}, is done via a maximal matching algorithm, where an edge between an entity $e$ an an entity variation $v_e$ means $e \in v_e$. 
For an EnvironmentData-Variation $v_{env}$ that defines a Collection-RangeVariation of type $A$, see \ref{ssec:variations}, computing the Comparison between an EnvironmentData $env$ and this target $v_{env}$ leads to a list of reasons for each entity $e_{env}$ in the entity collection of $env$, why $e_{env} \not\in v, \forall v \in A$. This can be seen in Figure \ref{fig:experiment_description}, where for each entity of \textit{LiquidContainer} concept in the environment, there is a list of differences, i.e. Comparisons, created for why the respective entity does not match the defined variation on the top-right.

The second step of the procedure is to turn the list of differences into a list of \actions\ that can fix them. In notation, \action\ $A_x$ solves a difference in the concept property $p_x$. The system knows which properties \actions\ modify by analyzing the definition of their effects. Thus, \actions\ are created (parametrized) to fix the differences in entity properties.

% Because multiple instances can fit the instance variation, the third step is to match instances with the variations. Our matching optimization criterion is to minimize the amount of \textit{Actions} needed to fix the instances' property differences. \todo{continue!}

In the third step, each \action\ $A_x$ is converted into an execution plan $P_x$ that implements solving the difference $\delta_{p_x}$ in the environment. It is also possible that there is no possibility to implement the \action\ $A_x$ in the environment; this is represented as an execution plan $P_x = \emptyset$. An execution plan $P_x$ is otherwise, in its simplest form, a set of \skill\ alternatives $\left\{S_y\right\}$, where the \skill\ $S_y$ implements the \action\ $A_x$. There is the case to consider that the \skill\ $S_y$ has preconditions that are not met. And so, before executing the skill $S_y$, a different execution plan $P_{S_y}$ has to be computed and executed to allow the \skill\ $S_y$ to solve the property difference $\delta_{p_x}$. It is also possible that one single \skill\  $S_y$ is not enough to implement the \action\ $A_x$. Consider the case where the environment contains three cups with $0.1L$ of water, and the goal is to have one cup with $0.3L$ of content. One single \textit{Pouring} \skill\ is not enough to fulfill the goal; two \textit{Pouring} \skills\ must be executed. Thus, in the most general form, an execution plan $P_x = \left[\left\{ S_{iy}, P_{S_{iy}} \right\}_i\right]$ is a list of skill alternatives $\left\{ S_{iy}, P_{S_{iy}} \right\}_i$, that possibly contain other execution plans $P_{S_{iy}}$ to solve the skill's preconditions.

Our procedure to parameterize the \skills\ $S_y$ that implement the \action\ $A_x$ is a custom solution for each property $p_x$. One could backtrack through all possible parameter values of all possible skills to create a general solution that works for all properties. Another idea is to invert \skill\ effects and thus guide the \skill\ parameter search from the target variation to the value. However, both approaches would be computationally intense and would not create execution plans in a reasonable time. 
% reinforcement learning with policy for each property

The procedure to solve an entity $e$'s \underline{contentLevel} property difference searches for other \textit{Container} object instances in the environment, sorts them according to their content volume, and iterates through them in ascending order if $e.contentLevel \le target.contentLevel$; otherwise, in descending order. If a \skill\ $S$ can be executed with the two objects, that reduces the difference between $e.contentLevel$ and $target.contentLevel$, the \skill\ is added to the execution plan. If, after checking all objects, $e.contentLevel \not\in target.contentLevel$, there is no solution to solve this property difference.

Thus, the result of the third step is an execution plan $P_x$ for each entity property difference.

Fourth, after having the execution plans $P_x$ per entity-variation and entity, a \underline{solution selector} scores all solutions according to defined metrics and then, via a maximal matching algorithm, selects the solutions to execute to satisfy all variations of the Collection-RangeVariation of type $A$. The edges in the maximal matching have the cost of the solution score. For this paper, the scoring metric by the \underline{solution selector} is the number of steps of the execution plan.

\begin{figure}[t!]
    % \vspace{-0.2cm}
    \centering
    \includegraphics[width=1\linewidth]{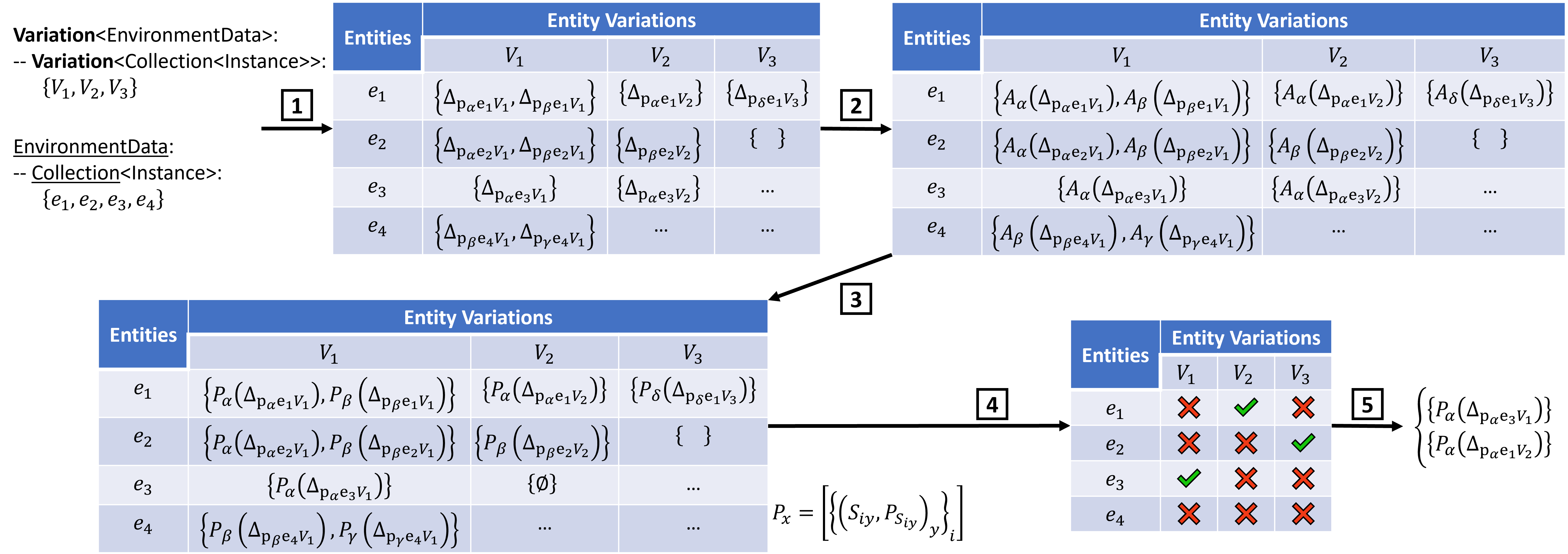}
    \caption{Data flow when transforming an environment into a given goal state. $\Delta$ are differences of entity properties $p$, $A$ are \actions, $P$ is an execution plan and $S$ are \skills.} \label{fig:data_flow}
\end{figure}

The fifth and final step is to pass the execution plan to the agent(s) to execute in the environment. Figure \ref{fig:data_flow} presents the flow of data through the five steps.
We have used the Franka Emika Panda robot in CoppeliaSim \cite{coppeliaSim} to perform the computed execution plan.
% Note that the approach is independent of the used robot; only when instantiating \skills\ must the robot's abilities, manipulability region, and workspace be considered. How the \skills\ are executed in the environment is separated from the modeling of what must be done.

\begin{figure}[t!]
    % \vspace{-0.2cm}
    \centering
    \includegraphics[width=1\linewidth]{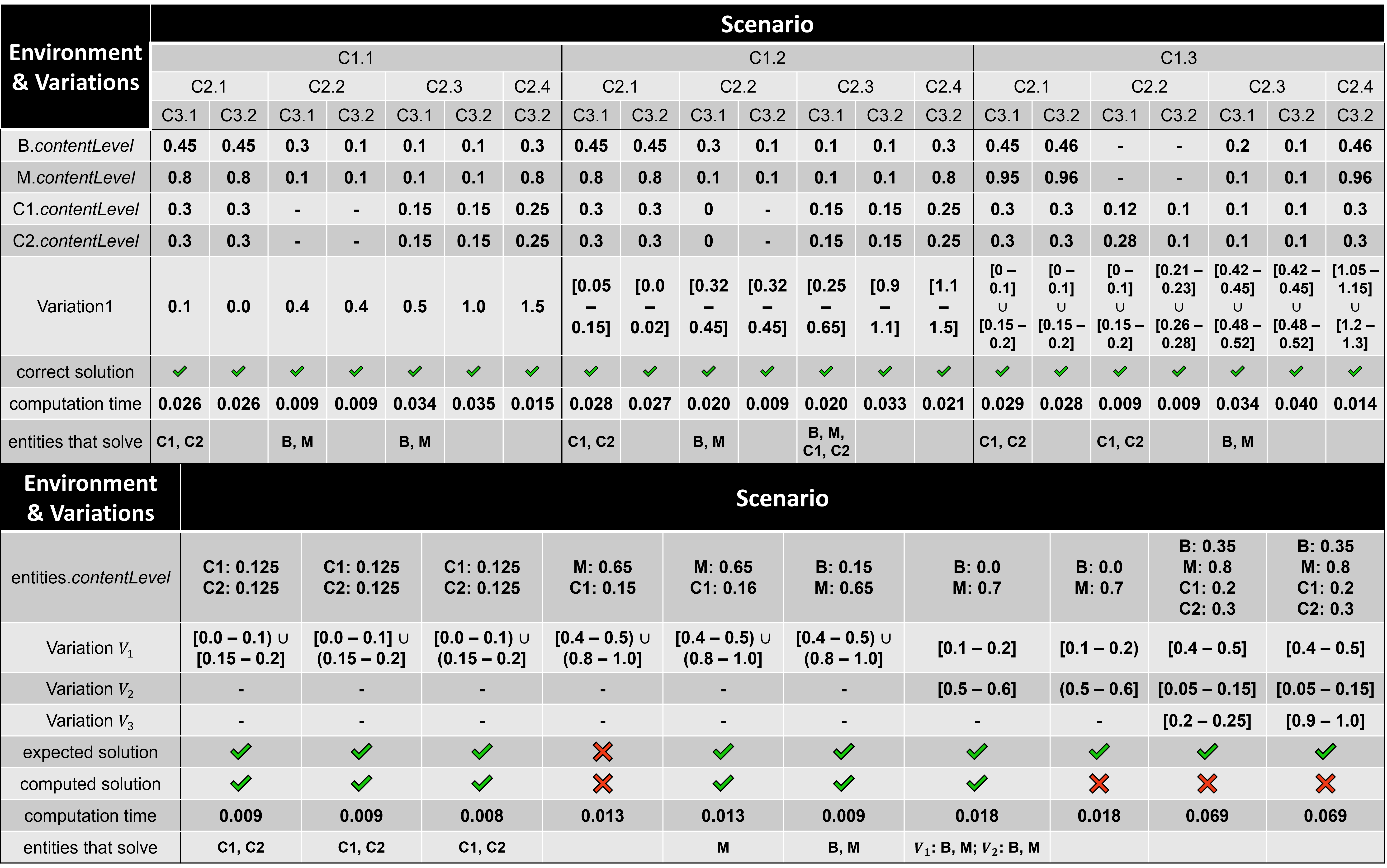}
    \caption{$B$ is a bowl with $0.5L$ \textit{contentVolume}, $M$ is a milk carton with $1.0L$ \textit{contentVolume}, $C1$ and $C2$ are cups with $0.3L$ \textit{contentVolume} each. Times, in seconds, averaged across 10 runs. Criteria $C2.4$ and $C3.1$ are mutually exclusive (a solution does not exist to let a container have more \textit{contentLevel} than its \textit{contentVolume}); thus, they are not included in the upper table. The lower table presents results for open intervals and multiple variations in the environment.} \label{fig:experiment_table}
\end{figure}

The experiments aim to compute solution plans for solving the difference of the \textbf{contentLevel} property of \textit{Container} objects. For this, we consider the following criteria. $C1$: \textbf{variation} type = $\left\{\text{fixed},\text{interval},\text{interval union}\right\}$. $C2$: target relative to content = \{$\left\{t < cL \le cV \right\}$, $\left\{cL < t < cV \right\}$, $\left\{cL < t \ni cV \right\}$, $\left\{cL \le cV < t \right\}$\}, where $t$ is the \textbf{variation} value and $cL$ and $cV$ are the \textit{contentLevel} and \textit{contentVolume} properties respectively. $C3$: achievable in environment $ = \left\{\text{yes}, \text{no}\right\}$. Figure \ref{fig:experiment_table} presents planning results for different environments and the criteria described above. The lower table shows cases where the computed solution does not match the actual solution. This only happens when multiple instance variations are defined. The reason is that the implemented procedure to turn the list of differences into an execution plan treats each difference independently. Thus, dependencies between two variations are not accurately solved.

\begin{figure}[t!]
    %\vspace{-0.1cm}
    \centering
    \includegraphics[width=1\linewidth]{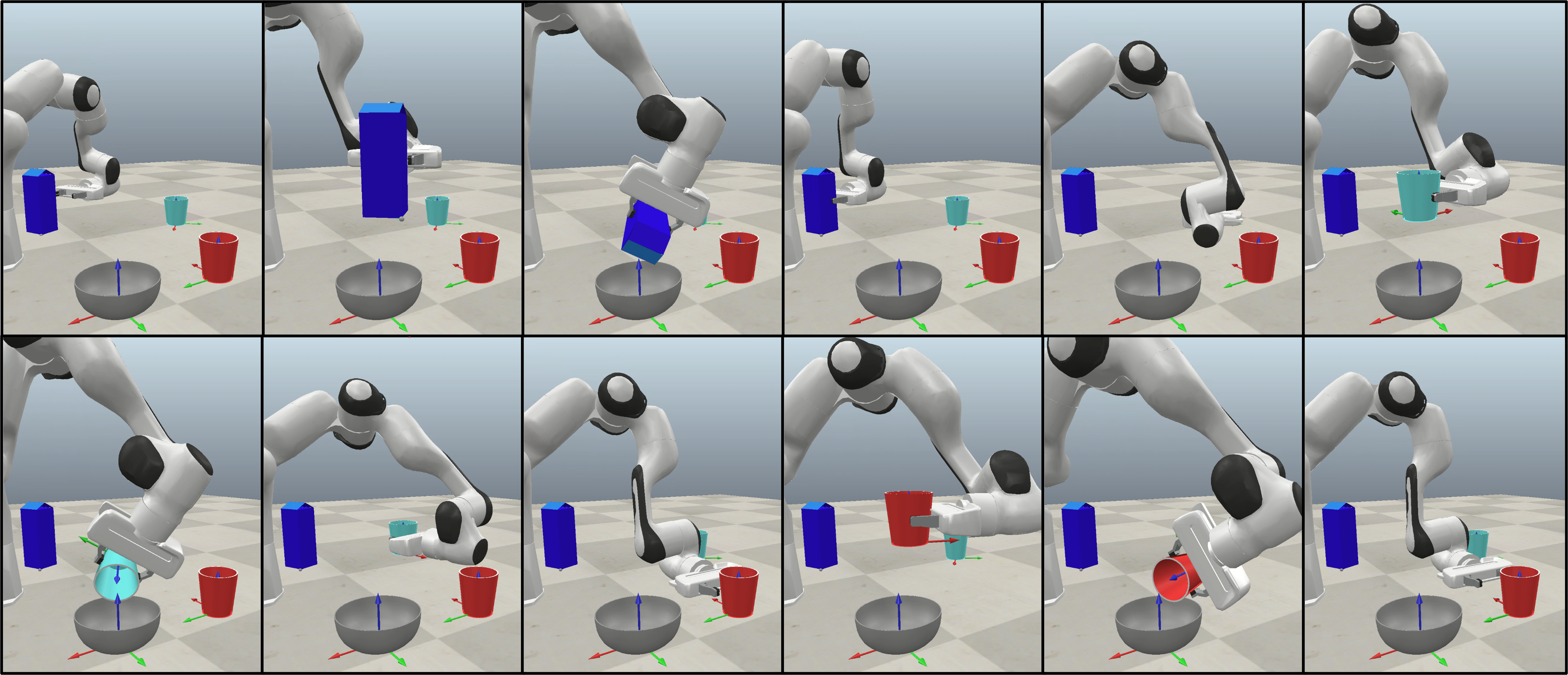}
    \caption{Robot executing plan to bring $B$, the bowl, into the goal state. Because no liquids were simulated, the pouring amount was associated with the pouring time via: $t_{pour} = 10 * amount_{pour}$.} \label{fig:robot_plan_execution}
\end{figure}

In the upper table of Figure \ref{fig:experiment_table}, there are two solutions for $C1.3$, $C2.3$, $C3.1$: one with the bowl $B$ as the instance in the \textbf{variation} $V1$, the other with $M$. The solution when $B$ is the matched instance has three steps: 1) pouring $0.1L$ from $M$ into $B$, 2) pouring $0.1L$ from $C1$ into $B$, and, finally, 3) pouring  $0.02L$ from $C2$ into $B$. This plan is sent to the robot in simulation and is executed as shown in Figure \ref{fig:robot_plan_execution}.

% \begin{figure}[t!]
%     % \vspace{-0.2cm}
%     \centering
%     \includegraphics[width=1\linewidth]{images/Experiment_Table_1Variation_compressed.png}
%     \caption{$B$ is a bowl with $0.5L$ \textit{contentVolume}, $M$ is a milk carton with $1.0L$ \textit{contentVolume}, $C1$ and $C2$ are cups with $0.3L$ \textit{contentVolume} each. Times, in seconds, averaged across 10 runs. Criteria $C2.4$ and $C3.1$ are mutually exclusive (a solution does not exist to let a container have more \textit{contentLevel} than its \textit{contentVolume}); thus, they are not included in the table.} \label{fig:experiment_table}
% \end{figure}

\section{CONCLUSIONS}
In this paper, we presented a model to represent the desired variation in a task's goal, i.e., the variation in an environment state. We also showed how to build this model from a single user demonstration of the task and how to use the model of the task's variation to create an execution plan to change a given environment into the goal state or determine if the goal state is unattainable with the \skills\ than an agent possesses.

With such a model, an agent would not need to imitate observed task execution trajectories, but could optimize the execution plan to its kinematic structure and attained \skills.

\subsection{Limitations}
The procedure detailed in \ref{ssec:exp_model_use} works for one instance variation. When multiple instance variations are defined in the environment variation, the procedure does not consider that to solve one instance variation, other instances will be modified. And thus, determined solution plans do not bring the whole environment into the goal state; just parts of it. This limitation is overcome by improving the planning procedure, e.g., by using a PDDL solver.

% The computation of execution plans treats the difference in entity properties $\delta_{p_x}$ as independent and, thus, there are cases where, e.g. entity $e_1$ fulfills the variation $v_1$ but for $e_2$ to fulfill variation $v_2$, a \skill\ $S_y$ is executed/selected, that modifies the property of $e_1$ which makes $e_1 \not\in v_1$ anymore.

Another limitation of the framework is the missing procedure to fix differences in \skill-preconditions, so that, e.g., if the milk carton is closed, the solution to open it and pour from it is determined by the system. The solution is to make the difference-solving procedure recursive and parameterize it with the differences found in \skill\ preconditions.

\subsection{Future Work}
Immediate future work is overcoming the limitations as presented above, followed by improving the scoring of \skills\ when selecting execution plans to consider the abilities of agents and, e.g., energy cost or path distance or time to completion of \skills. Furthermore, we plan to extend the variations to \textit{exclude} certain ranges of values. Including negations of ranges besides unions and intersections, would give the variations full expressiveness over the \textit{ValueDomains}.

\bibliographystyle{IEEEtran}
\bibliography{references}

\end{document}